\documentclass[preprint,12pt]{elsarticle}

\usepackage{amssymb}
\usepackage{times}
\usepackage{epsfig}
\usepackage{graphicx}
\usepackage{algorithm}
\usepackage{algorithmic}
\usepackage{amsmath}
\usepackage{amssymb}
\usepackage{mathrsfs}
\usepackage{url}
\usepackage{diagbox}
\usepackage{footmisc}
\usepackage{subfig}
\usepackage{float}
\usepackage{multirow}
\usepackage{array}
\usepackage{color}
\usepackage{booktabs}

\makeatletter

\newcommand{\Rmnum}[1]{\expandafter\@slowromancap\romannumeral #1@}
\makeatother

\newcommand{\tabincell}[2]{\begin{tabular}{@{}#1@{}}#2\end{tabular}}

\journal{arXiv}

\begin{document}

\begin{frontmatter}



\title{Part-based Visual Tracking via Structural Support Correlation Filter}


\author[label1,label2,label3]{Zhangjian Ji}
\author[label1,label2]{Kai Feng}
\author[label1,label2,label3]{Yuhua Qian}
\address[label1]{School of Computer \& Information Technology, Shanxi University, Taiyuan, China}
\address[label2]{ Key Laboratory of Computational Intelligence and Chinese Information Processing of Ministry of Education, Shanxi University, Taiyuan, China}
\address[label3]{ Institute of Big Data Science and Industry, Shanxi University, Taiyuan, China}
\begin{abstract}
   Recently, part-based and support vector machines (SVM) based trackers have shown favorable performance. Nonetheless, the time-consuming online training and updating process limit their real-time applications. In order to better deal with the partial occlusion issue and improve their efficiency, we propose a novel part-based structural support correlation filter tracking method, which absorbs the strong discriminative ability from SVM and the excellent property of part-based tracking methods which is less sensitive to partial occlusion. Then, our proposed model can learn the support correlation filter of each part jointly by a star structure model, which preserves the spatial layout structure among parts and tolerates outliers of parts. In addition, to mitigate the issue of drift away from object further, we introduce inter-frame consistencies of local parts into our model. Finally, in our model, we accurately estimate the scale changes of object by the relative distance change among reliable parts. The extensive empirical evaluations on three benchmark datasets: OTB2015, TempleColor128 and VOT2015 demonstrate that the proposed method performs superiorly against several state-of-the-art trackers in terms of tracking accuracy, speed and robustness.
\end{abstract}


\end{frontmatter}


\section{Introduction}
%
%

%
%
%
%
Visual object tracking has been an important research topic in the computer vision field and has a wide range of practical applications, \emph{e.g.}, intelligent surveillance, autonomous navigation of vehicles, human computer interaction, action recognition. Although great progress has been made in the past decades, it is still a challenging problem to design a robust visual tracking algorithm for real scenes, due to some complex situations, \emph{e.g.}, partial occlusion, illumination variation, pose changes, background clutter, complex motion and object blur. Here, we mainly investigate the key problem of learning a robust tracking model under these conditions mentioned above.

As is known, the discriminative models \cite{Authors36,Authors37,Authors38,Authors22,Authors16} have better performance than the generative models \cite{Authors12,Authors14,Authors15,Authors11,Authors9} in visual tracking. They seek to design a robust classifier to detect the target, and establish an optimal mechanism to update the model at each frame. For example, in order to realize the visual tracking, Avidan \cite{Authors38} adopted the SVM as an off-line binary classifier to detect target at each frame. Hare \emph{et al}. \cite{Authors37} applied the SVM with structured output to tracking the target because of its success in object detection. Although these two methods obtain the good results in visual tracking, the complex optimization still brings them the high computational complexity, which would make them not meet real-time applications, especially when considering the scale change of target and increasing feature dimensions of target representation. Recently, correlation filter (CF) utilizing the circulant property of dense sampling of base sample has attracted extensive attention in visual tracking due to its significant computational efficiency and robustness. Nevertheless, how to exploit the circulant property to accelerate SVM-based trackers remains unaddressed. Later, in view of the success of the max-margin CF (MMCF) \cite{Authors44} in the localization and classification of image, Zuo \emph{et al.} \cite{Authors45} developed the novel discriminative tracking algorithms based on support correlation filters that perform efficiently and accurately. Although obtained competitive results both in accuracy and robustness, all these methods are sensitive to the occlusion or partial occlusion.

To deal with the above issues, deformable part-based tracking methods \cite{Authors21, Authors46, Authors47,Authors48,Authors54} become more popular partially because of their favorable property of robustness against partial occlusion. Yao \emph{et al.} \cite{Authors54} employed an online structured output learning with latent variables to learn the weight parameters for an object and its parts, and distinguish the target object from the background using the weight parameters consequently. But their method fails to resolve the high computational complexity of the SVM. The researchers in \cite{Authors46, Authors47} brought the correlation filter into the part-based tracking methods which improves the tracking efficiency and robustness. However, their approaches ignore the spatial relations among object parts. More recently, Liu \emph{et al.} \cite{Authors48} improved the performance of their tracker by introducing structural constraints among parts into correlation filter. But they also don't consider the temporal consistency of motion model which would help to alleviate the problem of drift away from object.

Considering the existing problems of the methods mentioned above, in this paper, we build an efficient part-based support vector correlation filter tracking algorithm which is able to deal with partial occlusion and deformation effectively. Our method adopts the support vector correlation filer as the classifier of each part which absorbs strong discriminative ablility from SVM and speedups the SVM by the FFT in the Fourier domain. Then, our proposed model can learn the support correlation filter of each part jointly by a star structure model, which preserves the spatial layout structure among parts and tolerates outliers of parts. To further enhance the robustness of our model, we take into consideration the temporal consistency of each part, and incorporate it into our model to mitigate the issue of drift away from object. In addition, in order to adapt our tracker to scale changes of tracked target, we estimate the scale changes of object by the relative distance changes of the reliable part pairs. Finally, different from other multi-part trackers, we only estimate the position of the whole object by the tracking results of all visible parts, where each part is distinguished whether to be occluded by the PSR and appearance similarity.
\section{Related Work}
In this section, we only introduce the methods closely related to this work: SVM-based trackers, correlation filter trackers and part-based trackers in detail. For a survey of more tracking methods, we refer the reader to \cite{Authors5,Authors6}.

\textbf{SVM-based tracker:} Babenko \emph{et al.} \cite{Authors36} employed an online Multiple Instance Learning based appearance model to resolve the sample ambiguity problem. Hare \emph{et al.} \cite{Authors37} used the structure SVM with kernels to track the whole target. In \cite{Authors42}, an explicit feature mapping function is used to approximate nonlinear kernels. However, the complex optimization of SVM still brings them the high computational complexity, which prevents them from using the higher dimensional features. In 2013, Henriques \emph{et al.} \cite{Authors43} first applied the circulant property for training of support vector regression efficiently to detect pedestrians. Inspired by this work, Zuo \emph{et al.} \cite{Authors45} adopted the circulant property to design the support correlation filter tracker that perform efficiently and accurately, which lower the computational complexity $\mathcal{O}(n^{4})$ of SVM based trackers to $\mathcal{O}(n^{2}\log(n))$ for an $n\times n$ image patch. Wang \emph{et al.} \cite{Authors55} proposed a novel structured SVM based tracking method which takes dense circular samples into account in both training and detection processes.

\textbf{Correlation filter trackers:}  Bolme \emph{et al.} \cite{Authors22} first introduced the correlation filter into the visual tracking field because it can achieve the appealing results in terms of accuracy, robustness and speed. Afterwards, Henriques \emph{et al}. \cite{Authors27} incorporated multi-channel features into their kernelized correlation filters (KCF) framework to improve the accuracy and robustness of the tracker. However, they are only limited to estimate the target translation and signify poor performance when the targets of sequences involve significant scale variations. Thus, in order to adapt to the scale changes of the tracked target, Montero \emph{et al}. \cite{Authors58} use a similar approach (scale ratios between matched relevant keypoints) as in TLD \cite{Authors59} and \cite{Authors60} to estimate the size of tracked target. Danelljan \emph{et al}. \cite{Authors23} proposed a separable scale filter based on a scale pyramid representation to estimate the scale variation of target. And Li \emph{et al.} \cite{Authors24} adopted a multiple scales searching strategy to surmount the limitation that the conventional correlation filter (CF) trackers can not handle the scale variation of tracked target. Although the traditional correlation filter has obtained great success, unwanted boundary effects produced by the Fast Fourier Transform (FFT) result in an inaccurate description of the image, which will severely degrade the discriminative power of the learned model. To resolve this issue, Galoogahi \emph{et al}. \cite{Authors25} chose a larger searching size and then cropped the central patch of the signal that is same as the size of the filter by the binary matrix $\textbf{P}$ in each Alternating Direction Method of Multipliers (ADMM) iteration. Danelljan \emph{et al.} \cite{Authors26} utilized a spatially regularized component to deal with the boundary effect caused by the FFT, which achieves better tracking accuracy.

\textbf{part-based tracker:} To deal with the occlusion, many part-based trackers divided the entire target into separate parts \cite{Authors18,Authors46,Authors47,Authors21,Authors48,Authors51}. Liu \emph{et al.} \cite{Authors47} adapted the correlation filter as part classifiers. Akin \emph{et al.} \cite{Authors51} proposed a deformable part-based correlation filter tracking approach which depends on coupled interactions between a global filter and several part filters. Luke{\v z}i{\v c} \emph{et al.} \cite{Authors56} presented a new class of layered part-based trackers that apply a geometrically constrained constellation of local correlation filters for object localization.


\section{Structural Support Correlation Filter Tracker}\label{algsection}
In this section, we present an efficient part-based support vector correlation filter tracking algorithm. Since the proposed approach works in the framework of support correlation filter, we first briefly review the theory of support correlation filter in subsection \ref{svcf}. Then, in subsection \ref{KSCF}, we deduce the support correlation filter model in nonlinear space. Subsequently, in subsection \ref{FSSCF}, we give a detailed description of our proposed part-based structural support correlation filter tracker. Next the detailed solving procedures of our tracking approach are deduced in subsection \ref{optimization}. Finally, in subsection \ref{tracking}, we introduce a valid method that estimates the scale changes of object. Meanwhile, we also present a model update strategy by using the feedback from tracking results to avoid the model corruption.

In order to make our paper more readable, we first define some generic notations that will be useful before deriving our model, which is shown in Table \ref{Tb:5}.
\begin{table}[!h]
\caption{Define some generic notations which will be used in our work.}
\begin{center}
\begin{tabular}{cc}
  \toprule
  \textbf{Notation} & \textbf{Explanation }\\
  \hline
  $M, N$  & defined two given positive integers\\
  $\mathbb{R}$ & The set of real numbers\\
  $\hat{\textbf{u}}$ & The Fourier coefficients of $\textbf{u}$, $\forall \ \textbf{u}\in \mathbb{R}^{MN} $\\
  $\hat{\textbf{u}}^{*}$ & The complex conjugate of the Fourier coefficients of $\textbf{u}$\\
  $\mathcal{F}(\bullet)$ & The Fourier transform \\
  $\mathcal{F}^{-1}(\bullet)$& The inverse of $\mathcal{F}$\\
  $F$  & The base vectors of the discrete Fourier transform\\
  $F^{H}$ & The Hermitian transpose of $F$\\
  $\circ $ & \tabincell{c}{Indicated the element-wise multiplication of\\ any two vectors}\\
  $\widetilde{\max}\{\bullet,0\}$&\tabincell{c}{Calculated the maximum value between \\each elementof any vector and the zero}\\
  $\textbf{e}$& \tabincell{c}{defined an $MN\times 1$ vector,\\ each element of which is 1} \\
  $\textbf{E}$& \tabincell{c}{defined an $MN\times MN$ matrix, \\each element of which is 1} \\
  \bottomrule
\end{tabular}
\end{center}
\label{Tb:5}
\end{table}
\subsection{Review of Support Correlation Filter}\label{svcf}
Given a vectorized image patch $\textbf{x}\in \mathbb{R}^{MN}$, Zuo \emph{et al.} \cite{Authors45} learn a support correlation filter $\textbf{w}$ and a bias $b$ to classify any circular shift image $\textbf{x}_{m,n}$ of $\textbf{x}$  by
\begin{equation}\label{eq1}
  y_{m,n}=sgn(\textbf{w}^{T}\textbf{x}_{m,n}+b),
\end{equation}
Note that $m\in\{0,1,\cdots, M-1\}$ and $n\in\{0,1,\cdots, N-1\}$. $y_{m,n}$ denotes corresponding class label of one possible observation $\textbf{x}_{m,n}$ of a target object and all circular shift image $\textbf{x}_{m,n}$ forms a circulant matrix $\textbf{X}$. In general, $\textbf{X}$ can be expressed as
\begin{equation}\label{eq2}
  \textbf{X}=F^{H}diag(\hat{\textbf{x}})F,
\end{equation}
Then, classify all the samples of $\textbf{X}$ by
\begin{equation}\label{eq3}
  \textbf{y}=sgn(\mathcal{F}^{-1}(\hat{\textbf{x}}^{*}\circ \hat{\textbf{w}})+b\textbf{e}),
\end{equation}

Given the training sample set $\textbf{X}$ that consists of all circular shift image $\textbf{x}_{m,n}$ and its corresponding class label $\textbf{y}=[y_{0,0},\cdots,y_{m,n},\cdots,y_{M-1,N-1}]^{T}$, they use the squared hinge loss to define the SVM model \cite{Authors49} as follows:
\begin{equation}\label{eq4}
\begin{split}
  &\min_{\textbf{w},b}||\textbf{w}||^{2}+C||\boldsymbol{\xi}||^{2} \\
  &s.t. \ \ \textbf{y}\circ(\textbf{X}\textbf{w}^{T}+b\textbf{e})\geq \textbf{e}-\boldsymbol{\xi},
\end{split}
\end{equation}
where $\boldsymbol{\xi}=[\xi_{0,0},\cdots,\xi_{m,n},\cdots,\xi_{M-1,N-1}]^{T}$ is the vector of slack variables, $C$ is a trade-off parameter.

Based on the circulant property of $\textbf{X}$, the SVM model can be equivalently formulated as:
\begin{equation}\label{eq5}
\begin{split}
  &\min_{\textbf{w},b}||\textbf{w}||^{2}+C||\boldsymbol{\xi}||^{2} \\
  &s.t. \ \ \textbf{y}\circ(\mathcal{F}^{-1}(\hat{\textbf{x}}^{*}\circ\hat{\textbf{w}})+b\textbf{e})\geq \textbf{e}-\boldsymbol{\xi},
\end{split}
\end{equation}

In the SVM discriminative model, Zuo \emph{et al.} \cite{Authors45} assign binary class label by the confidence map of object position \cite{Authors3}, where the confidence map is defined as:
\begin{equation}\label{eq6}
  s(\textbf{p}_{m,n},\textbf{p}^{\star})=\Gamma exp(-\eta||\textbf{p}_{m,n}-\textbf{p}^{\star}||^\lambda),
\end{equation}
where $\textbf{p}^{\star}$ denotes the centre position of the interested object $\textbf{x}^{\star}$, $\textbf{p}_{m,n}$ represents the centre position of the circular shift image $\textbf{x}_{m,n}$, $\Gamma$ is a normalization constant, $\eta$ and $\lambda$ are the scale
and shape parameters respectively. With the confidence map, the class label $\textbf{y}$ can be obtained by
\begin{eqnarray}\label{eq7}
    y_{m,n}=\left\{
    \begin{array}{c}
      \begin{split}
      & 1  \quad if\; s(\textbf{p}_{m,n},\textbf{p}^{\star})\geq\theta_{u}\\
      -&1 \quad if\; s(\textbf{p}_{m,n},\textbf{p}^{\star})\leq\theta_{l}\\
      & 0  \quad otherwise
      \end{split}
    \end{array},
    \right.
\end{eqnarray}
where $\theta_{l}$ and $\theta_{u}$ are lower and upper thresholds respectively.

In order to exploit the property of the circulant matrix to learn the model (\ref{eq5}), let $\boldsymbol{\xi}=\textbf{v}+\textbf{e}-\textbf{y}\circ(\mathcal{F}^{-1}(\hat{\textbf{w}}\circ\hat{\textbf{x}}^{*})+b\textbf{e})$, and then it can be rewritten as:
\begin{equation}\label{eq8}
\begin{split}
  &\min_{\textbf{w},b,\textbf{v}}||\textbf{w}||^{2}+C||\textbf{y}\circ(\mathcal{F}^{-1}(\hat{\textbf{x}}^{*}\circ\hat{\textbf{w}})+b\textbf{e})-\textbf{e}-\textbf{v}||^{2} \\
  &s.t. \ \ \textbf{v}\succeq 0,
\end{split}
\end{equation}
where $\textbf{v}$ is an auxiliary variable and $\succeq$ denotes that each element of $\textbf{v}$ is greater than or equal to zero.

\subsection{Support Correlation Filter in Nonlinear Space}\label{KSCF}
To make the support correlation filter (SCF) model to be extended to learn the nonlinear decision function, we now derive a ``dual version" for the SCF model. In this derivation we partially follow Vapnik \cite{Authors50}. We start with re-expressing the SVM model in (\ref{eq4}) as:
\begin{equation}\label{eq9}
\begin{split}
  &\min_{\textbf{w},b,\textbf{v},\boldsymbol{\alpha}}||\textbf{w}||^{2}+\boldsymbol{\alpha}^{T}(\textbf{e}+\textbf{v}-\textbf{y}\circ(\textbf{X}\textbf{w}^{T}+b\textbf{e})-\boldsymbol{\xi}) \\&+C||\boldsymbol{\xi}||^{2}
  \\
  &s.t. \ \ \textbf{v}\succeq 0,
\end{split}
\end{equation}
Here $\boldsymbol{\alpha}$ is the Lagrange multiplier (it also represents the solution of SCF in the dual space). We let $\textbf{q}=\textbf{y}+\textbf{y}\circ \textbf{v}$, where $\textbf{v}\succeq 0$, and then the model (\ref{eq9}) can be rewritten as:
\begin{equation}\label{eq10}
\begin{split}
  &\min_{\textbf{w},b,\textbf{q},\boldsymbol{\alpha},\boldsymbol{\xi}}||\textbf{w}||^{2}+\boldsymbol{\alpha}^{T}(\textbf{q}-(\textbf{X}\textbf{w}^{T}+b\textbf{e})-\textbf{y}\circ\boldsymbol{\xi}) \\&+C||\boldsymbol{\xi}||^{2}.
\end{split}
\end{equation}
Solving the model (\ref{eq10}) with respect to $\textbf{w}$, we can obtain $\textbf{w}=\frac{1}{2}\boldsymbol{\alpha}^{T}\textbf{X}$. Then Substituting this into (\ref{eq10}), we obtain
\begin{equation}\label{eq11}
\begin{split}
&\min_{\boldsymbol{\xi},b,\textbf{q},\boldsymbol{\alpha}}-\frac{1}{4}\boldsymbol{\alpha}^{T}\textbf{X}\textbf{X}^{T}\boldsymbol{\alpha}+\boldsymbol{\alpha}^{T}(\textbf{q}-b\textbf{e})-\boldsymbol{\alpha}^{T}(\textbf{y}\circ\boldsymbol{\xi})
\\&+C||\boldsymbol{\xi}||^{2}.
\\
\end{split}
\end{equation}
Calculating (\ref{eq11}) with respect to $\boldsymbol{\xi}$, we obtain $\boldsymbol{\xi}=\frac{1}{2C}\textbf{y}^{T}\boldsymbol{\alpha}$. Then Substituting this into (\ref{eq11}), we get
\begin{equation}\label{eq12}
\begin{split}
&\min_{b,\textbf{q},\boldsymbol{\alpha}}-\frac{1}{4}\boldsymbol{\alpha}^{T}\textbf{X}\textbf{X}^{T}\boldsymbol{\alpha}+\boldsymbol{\alpha}^{T}(\textbf{q}-b\textbf{e})-\frac{1}{4C}\boldsymbol{\alpha}^{T}\boldsymbol{\alpha},
\end{split}
\end{equation}

Thus, the closed form solution to our sub-problem on $\boldsymbol{\alpha}$ can be formulated as
\begin{equation}\label{eq13}
  \boldsymbol{\alpha}=\frac{1}{4}(\textbf{X}\textbf{X}^{T}+\frac{1}{C}\textbf{E})^{-1}(\textbf{q}-b\textbf{e}).
\end{equation}

Given a non-linear mapping function $\varphi(\textbf{x})$, we define $\emph{K}(\textbf{x},\textbf{x}^{'})=\langle\varphi(\textbf{x}),\varphi(\textbf{x}^{'})\rangle$, which can be used by some kernel function (e.g., Gaussian RBF and polynomial) with permutation invariant. Based on the circulant property of $\textbf{X}$, $\textbf{X}\textbf{X}^{T}$ can be represented as
\begin{equation}\label{eq14}
  \textbf{X}\textbf{X}^{T}=F^{H}diag(\hat{\textbf{x}}\circ\hat{\textbf{x}}^{*})F,
\end{equation}
Then, introducing non-linear feature mapping $\varphi(\textbf{x})$ into the formula (\ref{eq14}), it can be revised as
\begin{equation}\label{eq15}
  F^{H}diag(\varphi(\hat{\textbf{x}})\circ \varphi(\hat{\textbf{x}}^{*})F=F^{H}\hat{\textbf{k}}^{\textbf{x}\textbf{x}}F=\textbf{K},
\end{equation}
where $\hat{\textbf{k}}^{\textbf{x}\textbf{x}}$ is the Fourier transform of $K(\textbf{x},\textbf{x})$ and $\textbf{K}$ is a circulant kernel matrix.

Thus, the solution to the sub-problem of the kernelized support correlation filter on $\boldsymbol{\alpha}$ can be formulated as
\begin{equation}\label{eq16}
  \boldsymbol{\alpha}=\frac{1}{4}(\textbf{K}+\frac{1}{C}\textbf{E})^{-1}(\textbf{q}-b\textbf{e}).
\end{equation}
\subsection{Formulation of Structural Support Correlation Filter}\label{FSSCF}
The support correlation filter model mentioned above is only to learn a holistic appearance model, which is not robust for partial occlusion. In order to tackle this issue, we introduce part-based tracking strategy to the support correlation filter model. Given a target object, it is divided into $L$ non-overlapping parts with $M\times N$ pixels. Then, we can learn the dual optimization variable $\boldsymbol{\alpha}_{l}$ of support correlation filter $\textbf{w}_{l}$ of each part via (\ref{eq17})
\begin{equation}\label{eq17}
\begin{split}
\min_{b,\textbf{q}_{l},\boldsymbol{\alpha}_{l}}\sum_{l=1}^{L}&-\frac{1}{4}\boldsymbol{\alpha}_{l}^{T}\textbf{X}_{l}\textbf{X}_{l}^{T}\boldsymbol{\alpha}_{l}+\boldsymbol{\alpha}_{l}^{T}(\textbf{q}_{l}-b_{l}\textbf{e})\\
&-\frac{1}{4C}\boldsymbol{\alpha}_{l}^{T}\boldsymbol{\alpha}_{l},
\end{split}
\end{equation}
Here $\textbf{q}_{l}=\textbf{y}+\textbf{y}\circ \textbf{v}_{l}$, where $\textbf{v}_{l}$ is an auxiliary variable corresponding to the $l$th part. The $b_{l}$ corresponds to the bias of the $l$th part in the model and the $\textbf{X}_{l}$ consists of all circular shift image $\textbf{x}_{m,n}$ of the $l$th part, where $l=1,\cdots,L$.


Intuitively, the motion model of each local part should be close to each other to cover the entire target. In order to characterize the similar motion among local parts and tolerate slight discrepancy among them, we introduce a customized Laplacian regularization term in the model (\ref{eq17}), that is
\begin{equation}\label{eq18}
\begin{split}
\min_{b,\textbf{q}_{l},\boldsymbol{\alpha}_{l}}\sum_{l=1}^{L}&-\frac{1}{4}\boldsymbol{\alpha}_{l}^{T}\textbf{X}_{l}\textbf{X}_{l}^{T}\boldsymbol{\alpha}_{l}+\boldsymbol{\alpha}_{l}^{T}(\textbf{q}_{l}-b_{l}\textbf{e})\\
&-\frac{1}{4C}\boldsymbol{\alpha}_{l}^{T}\boldsymbol{\alpha}_{l}+\frac{\delta}{2}\sum_{i,j}||\boldsymbol{\alpha}_{i}-\boldsymbol{\alpha}_{j}||^{2}\omega_{i,j}\\
& \forall \ \ i,j\in L \ \ and \ \ i\neq j,
\end{split}
\end{equation}
where $\delta$ is the weight parameter of the Laplacian regularization term. $\omega_{i,j}$ denotes a penalty parameter whether two parts $i$ and $j$ have similar motion. If $\omega_{i,j}$ is larger, the motion of two parts is more consistent, vice versa.

However, this fully connected structure makes it intractable to solve the model in (\ref{eq18}). Thus, under the situation of hardly lowering the performance of the model in (\ref{eq18}), we simplify the connected structure of the model in (\ref{eq18}) by a star model. In the star model, each local part is connected by an edge with a dummy part $\textbf{x}_{r}$ which can be represented by the mean image of all the local parts, \emph{i.e.} there are no direct relation between any two parts. Thus, this requires a minor adaptation of the model in (\ref{eq18}), that is
\begin{equation}\label{eq19}
\begin{split}
\min_{b,\textbf{q}_{l},\boldsymbol{\alpha}_{l}}\sum_{l=1}^{L}&-\frac{1}{4}\boldsymbol{\alpha}_{l}^{T}\textbf{X}_{l}\textbf{X}_{l}^{T}\boldsymbol{\alpha}_{l}+\boldsymbol{\alpha}_{l}^{T}(\textbf{q}_{l}-b_{l}\textbf{e})\\
&-\frac{1}{4C}\boldsymbol{\alpha}_{l}^{T}\boldsymbol{\alpha}_{l}+\frac{\delta}{2}\sum_{l=1}^{L}||\boldsymbol{\alpha}_{l}-\boldsymbol{\alpha}_{r}||^{2}\omega_{l,r},
\end{split}
\end{equation}
Here $\boldsymbol{\alpha}_{r}$ denotes dual optimization variable of support correlation filter $\textbf{w}_{r}$ of the dummy part $\textbf{x}_{r}$. Because the target moves smoothly between consecutive two frames, we can use the motion consistency among parts in $(t-1)$th frame to describe the motion relation among parts at the current frame. So, we define $\omega_{l,r}$ to decrease exponentially with the hyper-distance of support correlation filters $\textbf{w}_{l}$ and $\textbf{w}_{r}$ of the $l$th part $\textbf{x}_{l}$ and the dummy part $\textbf{x}_{r}$ in $(t-1)$th frame, i.e.,
\begin{equation}\label{eq20}
  \omega_{l,r}=\exp(-\frac{1}{2}\frac{||\textbf{w}_{l}^{t-1}-\textbf{w}_{r}^{t-1}||^{2}}{\kappa^{2}}),
\end{equation}
where $\kappa$ is a smooth factor.

In practice, according to the observation, we found that the appearance of tracked object changes smoothly over time.
 Thus the selected training samples should be similar in consecutive frames. That is to say, the corresponding $\boldsymbol{\alpha}_{l}^{t-1}$ of each local part in $(t-1)$th frame should be close to that in $t$th frame, which is called temporal consistency. Therefore, we may introduce temporal constrain term into the model (\ref{eq19}) and revise it as follows
\begin{equation}\label{eq21}
\begin{split}
&\min_{b,\textbf{q}_{l},\boldsymbol{\alpha}_{l}}\sum_{l=1}^{L}(-\frac{1}{4}\boldsymbol{\alpha}_{l}^{T}\textbf{X}_{l}\textbf{X}_{l}^{T}\boldsymbol{\alpha}_{l}+\boldsymbol{\alpha}_{l}^{T}(\textbf{q}_{l}-b_{l}\textbf{e})\\
&-\frac{1}{4C}\boldsymbol{\alpha}_{l}^{T}\boldsymbol{\alpha}_{l})+\frac{\delta}{2}\sum_{l=1}^{L}||\boldsymbol{\alpha}_{l}-\boldsymbol{\alpha}_{r}||^{2}\omega_{l,r}\\
&+\frac{\beta}{2}\sum_{l=1}^{L}||\boldsymbol{\alpha}_{l}^{t}-\boldsymbol{\alpha}_{l}^{t-1}||^{2},
\end{split}
\end{equation}
where $\beta$ is the controlling factor of the temporal constrain term.

Given non-linear mapping function $\varphi(\textbf{x})$ and the derivation of formulas (\ref{eq14}) and (\ref{eq15}), our model in (\ref{eq21}) can be extended to learn a kernelized structured support correlation filter model, i.e.
\begin{equation}\label{eq22}
\begin{split}
&\min_{b_{l},\textbf{q}_{l},\boldsymbol{\alpha}_{l}}\sum_{l=1}^{L}(-\frac{1}{4}\boldsymbol{\alpha}_{l}^{T}\textbf{K}_{l}\boldsymbol{\alpha}_{l}+\boldsymbol{\alpha}_{l}^{T}(\textbf{q}_{l}-b_{l}\textbf{e})\\
&-\frac{1}{4C}\boldsymbol{\alpha}_{l}^{T}\boldsymbol{\alpha}_{l})+\frac{\delta}{2}\sum_{l=1}^{L}||\boldsymbol{\alpha}_{l}-\boldsymbol{\alpha}_{r}||^{2}\omega_{l,r}\\
&+\frac{\beta}{2}\sum_{l=1}^{L}||\boldsymbol{\alpha}_{l}^{t}-\boldsymbol{\alpha}_{l}^{t-1}||^{2}.
\end{split}
\end{equation}
where $\textbf{K}_{l}$ is a circulant kernel matrix corresponding to the $l$th part.

According to the above points, our models in (\ref{eq21}) and (\ref{eq22}) can learn the support correlation filter parameters of all local parts jointly and distinguish the parts from the background. Our model is also resistant to partial occlusion. Besides, it has high efficiency and robustness.
\subsection{Optimization}\label{optimization}
In this subsection, we utilize the Alternating Direction Method of Multipliers (ADMM) method \cite{Authors29} to solve the optimization problem in (\ref{eq21}). When keeping other variables fixed, the ADMM method can iteratively update one of the variables $b_{l}, \textbf{q}_{l},\boldsymbol{\alpha}_{l},\boldsymbol{\alpha}_{r}$ by minimizing (\ref{eq21}), which can guarantee the convergence of our proposed model. Consequently, updating steps corresponding to all the variables are as follows\\
\textbf{Step 1: update} $\boldsymbol{\alpha}_{r}$ (with others fixed): $\boldsymbol{\alpha}_{r}$ can be updated by solving the following optimization problem
\begin{equation}\label{eq23}
 \boldsymbol{\alpha}_{r} =\arg\min_{\boldsymbol{\alpha}_{r}}\frac{\delta}{2}\sum_{l=1}^{L}||\boldsymbol{\alpha}_{l}-\boldsymbol{\alpha}_{r}||^{2}\omega_{l,r},
\end{equation}
and its solution is
\begin{equation}\label{eq24}
 \boldsymbol{\alpha}_{r}=\frac{1}{\sum_{l=1}^{L}\omega_{l,r}} \sum_{l=1}^{L}\omega_{l,r}\boldsymbol{\alpha}_{l}.
\end{equation}
\textbf{Step 2: update} $\textbf{q}_{l}$ (with others fixed): before computing $\textbf{q}_{l}$, we firstly need to calculate the variable $\textbf{v}_{l}$. Combining the models (\ref{eq8}) and (\ref{eq9}), the subproblem on $\textbf{v}_{l}$ becomes
\begin{equation}\label{eq25}
 \begin{split}
  \textbf{v}_{l} =&\arg\min_{\textbf{v}_{l}}||\textbf{v}_{l}-(\textbf{y}\circ(\textbf{X}_{l}\textbf{w}_{l}^{T}+b_{l}\textbf{e})-1)||^{2}.\\
  &s.t. \ \ \textbf{v}_{l}\succeq 0
  \end{split}
\end{equation}
Then, $\textbf{v}_{l}$ has the following closed form solution:
\begin{equation}\label{eq26}
  \textbf{v}_{l}=\widetilde{\max}\{\textbf{y}\circ(\textbf{X}_{l}\textbf{w}_{l}^{T}+b_{l}\textbf{e})-\textbf{e},0\}.
\end{equation}
In view of $\textbf{w}_{l}=\frac{1}{2}\boldsymbol{\alpha}_{r}^{T}\textbf{X}$, the formula (\ref{eq26}) can be modified as
\begin{equation}\label{eq27}
  \textbf{v}_{l}=\widetilde{\max}\{\textbf{y}\circ(\frac{1}{2}\textbf{X}_{l}\textbf{X}_{l}^{T}\boldsymbol{\alpha}_{l}+b_{l}\textbf{e})-\textbf{e},0\},
\end{equation}
When $\textbf{x}_{l}$ is mapped to the kernel feature space, the amended version of the formula (\ref{eq27}) is as follows
\begin{equation}\label{eq28}
  \textbf{v}_{l}=\widetilde{\max}\{\textbf{y}\circ(\frac{1}{2}\textbf{K}_{l}\boldsymbol{\alpha}_{l}+b_{l}\textbf{e})-\textbf{e},0\},
\end{equation}
Known $\textbf{v}_{l}$ from the aforementioned formulas (\ref{eq27}) or (\ref{eq28}), we can calculate $\textbf{q}_{l}$ by
\begin{equation}\label{eq29}
  \textbf{q}_{l}=\textbf{y}+\textbf{y}\circ \textbf{v}_{l}.
\end{equation}
\textbf{Step 3: update} $b_{l}$ (with others fixed): we exploit the method of solving the parameter $b$ in \cite{Authors45} to calculate the $b_{l}$, \emph{i.e}
\begin{equation}\label{eq30}
  b_{l}=\bar{q}_{l}.
\end{equation}
where $\bar{q}_{l}$ is the mean of $\textbf{q}_{l}$.\\
\textbf{Step 4: update} $\boldsymbol{\alpha}_{l}$ (with others fixed): The minimization problem (\ref{eq21}) with respect to $\{\boldsymbol{\alpha}_{l}\}_{l=1}^{L}$ can be decomposed into $L$ mutually independent subproblems. The $l$th subproblem to update $\boldsymbol{\alpha}_{l}$ can be equivalently re-expressed as
\begin{equation}\label{eq31}
\begin{split}
&\boldsymbol{\alpha}_{l}=\arg\min_{\boldsymbol{\alpha}_{l}}-\frac{1}{4}\boldsymbol{\alpha}_{l}^{T}\textbf{X}_{l}\textbf{X}_{l}^{T}\boldsymbol{\alpha}_{l}+\boldsymbol{\alpha}_{l}^{T}(\textbf{q}_{l}-b_{l}\textbf{e})\\
&-\frac{1}{4C}\boldsymbol{\alpha}_{l}^{T}\boldsymbol{\alpha}_{l}+\frac{\delta}{2}||\boldsymbol{\alpha}_{l}-\boldsymbol{\alpha}_{r}||^{2}\omega_{l,r}\\
&+\frac{\beta}{2}||\boldsymbol{\alpha}_{l}^{t}-\boldsymbol{\alpha}_{l}^{t-1}||^{2},
\end{split}
\end{equation}
Then, for each $\boldsymbol{\alpha}_{l}$, the closed form solution of the formula (\ref{eq31}) is shown as follows
\begin{equation}\label{eq32}
\begin{split}
 \boldsymbol{\alpha}_{l}=&(\frac{1}{2}\textbf{X}_{l}\textbf{X}_{l}^{T}+\frac{1}{2C}\textbf{E}-\delta\omega_{l,r}\textbf{E}-\beta\textbf{E})^{-1}((\textbf{q}_{l}-b_{l}\textbf{e})\\
 &-\delta\omega_{l,r}\boldsymbol{\alpha}_{r}-\beta\boldsymbol{\alpha}_{l}^{t-1}).
\end{split}
\end{equation}

The detailed ADMM algorithm that solves our model (\ref{eq21}) is given in Algorithm \ref{Alg:1}, where the convergence is reached when the change of solution $\boldsymbol{\alpha}_{l}$ is below a pre-defined threshold (e.g. $\tau=10^{-3}$ in our work) or the number of iteration is greater than the maximum iterations $Iter$.

\begin{algorithm}
\hspace*{0.02in}\textbf{Input:} Training data: $\textbf{X}_{l}$ and $\textbf{y}$. Initialization of parameters $\delta,\beta,C$\\
\textbf{Output:} $\{\boldsymbol{\alpha}_{l},b_{l}\}_{l=1}^{L}$
\begin{algorithmic}[1]
\STATE Initialize $num\leftarrow1$, ${\boldsymbol{\alpha}_{l}^{t}}^{(1)}=\frac{\textbf{X}_{l}^{T}\textbf{y}}{\textbf{X}_{l}\textbf{X}_{l}^{T}+\frac{1}{C}\textbf{E}}$, ${\boldsymbol{\alpha}_{l}^{t}}^{(0)}=\textbf{0}$, $b_{l}^{t}=\bar{y}$, where $\bar{y}$ is the mean of $\textbf{y}$. \\

\WHILE{$num\leq Iter$ or $|{\boldsymbol{\alpha}_{l}^{t}}^{(i)}-{\boldsymbol{\alpha}_{l}^{t}}^{(i-1)}|>\tau$} 
{
\STATE  Update $\boldsymbol{\alpha}_{r}$ via (\ref{eq24})\\
\FOR{$l=1$ to $L$}
\STATE Update $\textbf{q}_{l}$ via (\ref{eq27}) and (\ref{eq29})
\STATE Update $b_{l}$ via (\ref{eq30})
\STATE  calculate $\boldsymbol{\alpha}_{l}^{t}$ via (\ref{eq32})
\ENDFOR
\STATE  $num\leftarrow num+1$
}
\ENDWHILE
\end{algorithmic}
\caption{Solving the optimization problem defined by the model (\ref{eq21})}
\label{Alg:1}
\end{algorithm}
As shown in algorithm \ref{Alg:1}, its major computing cost is that we need to calculate the matrix inverse and multiplication in spatial
domain when updating $\boldsymbol{\alpha}_{l}^{t}$ via (\ref{eq32}). However, in view of the circulant structure property of $\textbf{X}_{l}$, $\boldsymbol{\alpha}_{l}^{t}$ can be calculated very efficiently in the Fourier domain. Thus, the formula (\ref{eq32}) can be rewritten as the version (\ref{eq33}) in the Fourier domain.
\begin{equation}\label{eq33}
  \hat{\boldsymbol{\alpha}}_{l}^{t}=\frac{\hat{\textbf{q}}_{l}-b_{l}\hat{\textbf{e}}-\delta\omega_{l,r}\hat{\boldsymbol{\alpha}}_{r}-\beta\hat{\boldsymbol{\alpha}}_{l}^{t-1}}{\frac{1}{2}\hat{\textbf{x}}_{l}\circ\hat{\textbf{x}}_{l}^{*}+\frac{1}{2C}\hat{\textbf{e}}-\delta\omega_{l,r}\hat{\textbf{e}}-\beta\hat{\textbf{e}}},
\end{equation}
where $\frac{\bullet}{\bullet}$ denotes the element-wise division.

When the sample $\textbf{x}_{l}$ is mapped to the kernel feature space, the updating step with respect to $\boldsymbol{\alpha}_{l}$ needs a minor modification, that is
\begin{equation}\label{eq34}
\begin{split}
 \boldsymbol{\alpha}_{l}=&(\frac{1}{2}\textbf{K}_{l}+\frac{1}{2C}\textbf{E}-\delta\omega_{l,r}\textbf{E}-\beta\textbf{E})^{-1}((\textbf{q}_{l}-b_{l}\textbf{e})\\
 &-\delta\omega_{l,r}\boldsymbol{\alpha}_{r}-\beta\boldsymbol{\alpha}_{l}^{t-1}),
\end{split}
\end{equation}
Meanwhile, the corresponding version of the formula (\ref{eq34}) in the Fourier domain is as follows
\begin{equation}\label{eq35}
  \hat{\boldsymbol{\alpha}}_{l}^{t}=\frac{\hat{\textbf{q}}_{l}-b_{l}\hat{\textbf{e}}-\delta\omega_{l,r}\hat{\boldsymbol{\alpha}}_{r}-\beta\hat{\boldsymbol{\alpha}}_{l}^{t-1}}{\frac{1}{2}\hat{\textbf{k}}^{\textbf{x}\textbf{x}}+\frac{1}{2C}\hat{\textbf{e}}-\delta\omega_{l,r}\hat{\textbf{e}}-\beta\hat{\textbf{e}}}.
\end{equation}
where $\frac{\bullet}{\bullet}$ denotes the element-wise division.

To solve the optimization problem defined by (\ref{eq22}), we only need to use the formulas (\ref{eq28}) and (\ref{eq34}) to replace the formulas (\ref{eq27}) and (\ref{eq32}) in the updating steps.

Finally, known $\hat{\boldsymbol{\alpha}}_{l}$, the $\boldsymbol{\alpha}_{l}$ can be obtained by $\boldsymbol{\alpha}_{l}=\mathcal{F}^{-1}(\hat{\boldsymbol{\alpha}}_{l})$. Moreover, to speed up the algorithm \ref{Alg:1}, it can be implemented in matrix form without the ``for" loop.
\subsection{Tracking}\label{tracking}
At the tracking stage of nonlinear feature space, when obtaining the coefficient vector $\hat{\boldsymbol{\alpha}}_{l}^{t-1}$ and bias $b_{l}$ of each local part in the previous frame, we can estimate the response map of each local patch $\textbf{z}_{l}$ at the current frame by the following formula
\begin{equation}\label{eq36}
 \textbf{f}_{l}^{t}=\mathcal{F}^{-1}(\hat{\textbf{k}}_{l}^{\textbf{x}\textbf{z}}\circ\hat{\boldsymbol{\alpha}}_{l}^{t-1})+b_{l}\textbf{e},
\end{equation}
where $\hat{\textbf{k}}_{l}^{\textbf{x}\textbf{z}}$ denotes the Fourier transform of $K(\textbf{x},\textbf{z})$ for the $l$th part. The position $\textbf{p}_{l}^{t}$ of the $l$th part can be determined by the maximum value of $\textbf{f}_{l}^{t}$, e.g. the position $\textbf{p}_{l}^{t}$ of each part can be expressed as
\begin{equation}\label{eq37}
 \textbf{p}_{l}^{t} = \textbf{p}_{l}^{t-1}+\boldsymbol{\Delta}_{l}^{t},
\end{equation}
Here $\boldsymbol{\Delta}_{l}^{t}$ denotes the translation of the $l$th part at time $t$.

Then we can estimate the final position $\textbf{p}_{g}^{t}$ of object by the translation $\boldsymbol{\Delta}_{l}^{t}$ of all the parts, that is
\begin{equation}\label{eq38}
 \textbf{p}_{g}^{t}=\textbf{p}_{g}^{t-1} + \sum_{l=1}^{L}\pi_{l}\boldsymbol{\Delta}_{l}^{t},
\end{equation}
where $\pi_{l}$ is the weight parameter of corresponding part.

Because different parts of the target may suffer different appearance changes, illumination variation or occlusion in different frames, intuitively, if we assign the same weight to each part, the falsely tracker part may be overemphasized which will lead to drift problem. In order to handle this issue, we should adaptively give each part a different weight according to its reliability. In our work, we exploit the peak-to-sidelobe ratio (PSR) to define the weight of each part because the higher PSR usually means more reliable part, where the PSR is defined as
\begin{equation}\label{eq39}
  \phi_{l}=\frac{\max(\textbf{f}_{l})-\mu_{l}}{\sigma_{l}},
\end{equation}
where $\mu_{l}$ and $\sigma_{l}$ are the mean and the standard deviation of $\textbf{f}_{l}$ respectively.

In addition, for tracking problem, the appearance similarity between two consecutive frames is helpful for distinguishing whether the part is reliable or not. Taking this observation into consideration, we define the appearance similarity $d$ to determine whether the part is reliable, i.e.
\begin{equation}\label{eq40}
  d_{l}=exp(-\frac{||\textbf{x}_{l}^{t}-\textbf{x}_{l}^{t-1}||^{2}}{\gamma^{2}}),
\end{equation}
where $\gamma$ is a hyperparameter, and $\textbf{x}$ is the vector representation of object appearance features, where the appearance features use the color histogram features.
\begin{figure}[t]
   \centering
   \subfloat[]{\label{Fig:r1}
   \includegraphics[width=0.45\linewidth]{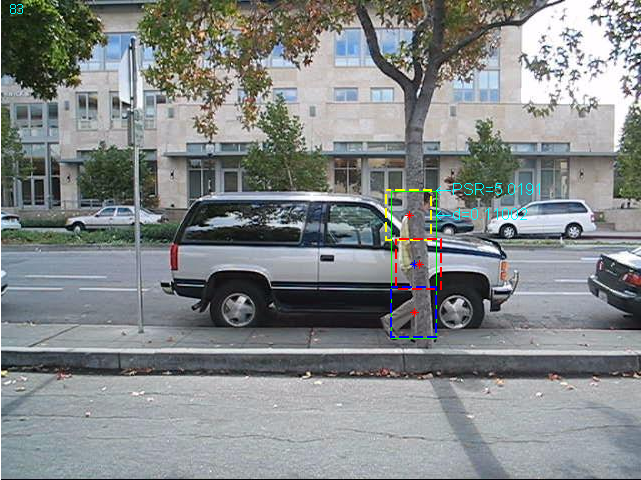}}
   \subfloat[]{\label{Fig:r2}
   \includegraphics[width=0.45\linewidth]{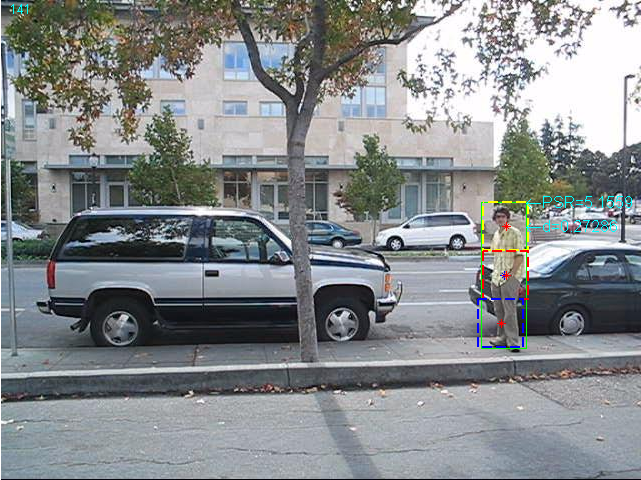}}\\

   \caption{Visualization of the PSR and appearance similarity of the part denoted by the yellow bounding box in the frame \#83 and \#141. (a) The target is occluded in the frame \#83. The $PSR=5.01191$ and the appearance similarity $d=0.11002$ of the part denoted by the yellow bounding box. (b) he target is occluded in the frame \#141. The $PSR=5.1559$ and the appearance similarity $d=0.27286$ of the part denoted by the yellow bounding box.}
\label{fig:occ}
\end{figure}
Combining two indicators above, we distinguish whether the $l$th part is occluded or has a large pose change. If $\phi_{l}$ and $d_{l}$ are less than the pre-defined threshold, this part is unreliable. As shown in Fig.\ref{fig:occ}(a), when the part that denoted by the yellow bounding box is occluded, its PSR and appearance similarity both become smaller. According to our two criteria, this part is unreliable. But in the Fig.\ref{fig:occ}(b), the target occurs the deformation. If using our two criteria, this part is reliable and if only using the PSR, then this part is unreliable. In fact, for the deformed part, we need to update its model to avoid drift.

To avoid erroneous estimations further, we use the PSR value and the appearance similarity $d_{l}$ from the reliable parts to calculate the corresponding weight $\pi_{j}$, i.e.
\begin{equation}\label{eq41}
  \pi_{j}=(1-\varpi)\frac{\phi_{j}}{\sum_{j=1}^{J}\phi_{j}}+\varpi\frac{d_{j}}{\Sigma_{j=1}^{J}d_{j}},
\end{equation}
where $J$ denotes the number of all reliable parts, $\varpi$ is a fusion parameter. In our work, $\varpi$ is set as 0.4. By now, the formula (\ref{eq38}) can be modified as
\begin{equation}\label{eq42}
 \textbf{p}_{g}^{t}=\textbf{p}_{g}^{t-1} + \sum_{j=1}^{J}\pi_{j}\boldsymbol{\Delta}_{j}^{t},
\end{equation}
Here if $J=0$, this means that all parts are unreliable. At this time, we use the translation of target in the previous frame to approximate that of target in the current frame because the motion of target hardly keep steady between two consecutive frames in most cases.\\
\textbf{Update scheme:} During tracking, the object appearance may change because of a number of challenging factors such as illumination change and pose change. Hence it is necessary to update part classifiers over time. Our tracking model is made up of the learned target appearance $\textbf{x}_{l}$ and the transformed classifier coefficients $\boldsymbol{\alpha}_{l}$. For each patch, our model parameters are updated by
\begin{equation}\label{eq43}
 \begin{split}
 &\textbf{x}_{l}^{t}=(1-\rho_{l})\textbf{x}_{l}^{t-1}+\rho_{l}\textbf{x}_{l}\\
 &\boldsymbol{\alpha}_{l}^{t}=(1-\rho_{l})\boldsymbol{\alpha}_{l}^{t-1}+\rho_{l}\boldsymbol{\alpha}_{l},
 \end{split}
\end{equation}
where $\rho_{l}$ is a learning rate parameter. The $\boldsymbol{\alpha}_{l}$ is calculated by simple linear interpolation. The $\textbf{x}_{l}$ is updated by taking the current appearance into account.

However, if using a fixed learning rate $\rho$ in the updating process, the whole model will be contaminated in
the remaining frames once the tracker loses the object. Thus, to avoid producing errors, It is apparent that the model of the occluded part should not be updated and other parts should adaptively adjust their learning rate based on the corresponding reliable weight. Therefore the learning rate of each part is updated by the following scheme
\begin{eqnarray}\label{eq44}
    \rho_{l}=\left\{
    \begin{array}{c}
      \begin{split}
      & \pi_{l}\varrho  \quad if\; \phi_{l}>\epsilon \ \ or \ \ d_{l}>\varepsilon\\
      & 0  \quad otherwise
      \end{split}
    \end{array},
    \right.
\end{eqnarray}
where $\varrho$ is a fixed learning rate, $\epsilon$ and $\varepsilon$ are two predefined thresholds.

Thus, contrary to traditional correlation filter based trackers, due to exploiting the adaptive updating strategy, our method can still maintain the tracking accuracy by using the results of the previous frame even when all part are occluded at one frame.\\
\textbf{Scaling:} To adapt the scale change of object, most of correlation filter based trackers \cite{Authors2,Authors23,Authors24} utilize a discriminative filter or a search pool that is based on pyramidal structure to estimate the object size. Despite obtaining outstanding results, these methods do not accurately estimate the current object size. So, to tackle this issue, we adopt the ration of the relative distance among local parts as in \cite{Authors51} to estimate the object size accurately because it's positively correlated with the scale of the target. In addition, to improve the accuracy of estimating object size further, in our work, we only use the change rate of relative distance among reliable local parts to estimate the object size. Therefore, the object scale $S^{t}$ is calculated by
\begin{equation}\label{eq45}
  S^{t}=\frac{S^{t-1}}{J(J-1)}\sum_{i=1}^{J}\sum_{j=1}^{J}\frac{||\textbf{p}_{i}^{t}-\textbf{p}_{j}^{t}||^{2}}{||\textbf{p}_{i}^{t-1}-\textbf{p}_{j}^{t-1}||^{2}}\ \ (i\neq j).
\end{equation}
where $\textbf{p}_{i}^{t}$ represents the position of part $i$ in the $t$th frame. Because at least two reliable parts can make the formula (\ref{eq45}) feasible, we keep the scale size of the preview frame unchange when only one part is available. In addition, the scale of the target does not change dramatically between two consecutive frames. To estimate the scale of target more robustly, we utilize the moving average to calculate the scale of target at the current frame.

So far, the theoretical part of the algorithm has been completely introduced above. For better comprehending our proposed method, it is summarized in Algorithm \ref{Alg:2}.
\begin{algorithm}
\textbf{Input}: Image frames $\{I_{t}\}_{1}^{T}$, initial object position $\textbf{p}_{g}^{1}$\\
\textbf{Output}: Target position of each frame $\{\textbf{p}_{g}^{t}\}_{2}^{T}$\\
\vspace{-0.4cm}
\begin{algorithmic}[1]
\STATE \textbf{repeat}
\STATE Calculate the position $\textbf{p}_{l}^{t-1}$ of each part based on the last target position $\textbf{p}_{g}^{t-1}$.
\STATE Crop an image patch $\textbf{x}_{l}$ from $I_{t}$ at the patch position $\textbf{p}_{l}^{t-1}$ of the $t-1$ time and extract the corresponding feature representation.
\STATE Calculate the filter coefficient $\boldsymbol{\alpha}_{l}$ and bias $b_{l}$ of each patch by the algorithm \ref{Alg:1}.
\STATE Detection the position $\textbf{p}_{l}^{t}$ of each patch via (\ref{eq36}).
\STATE Distinguish whether the $l$th part is reliable by $\phi_{l}$ and $d_{l}$.
\STATE Compute the target position $\textbf{p}_{g}^{t}$ at the current time via (\ref{eq42}).
\STATE Estimate the scale of the target via (\ref{eq45}).
\STATE Update learned target appearance $\textbf{x}_{l}$ and and the transformed classifier coefficients $\boldsymbol{\alpha}_{l}$ with the formulas (\ref{eq43}) and (\ref{eq44}).
\STATE \textbf{until} end of video sequence.
\end{algorithmic}
\caption{Scale structural support kernel correlation filter tracking algorithm (ScaleSSKCF)}
\label{Alg:2}
\end{algorithm}
\section{Experiments}
In the experimental part, we use the several benchmark datasets: TempleColor128\footnote{\label{fn:2}The sequences together with the ground-truth and matlab evaluation toolkit is available at: \url{http://www.dabi.temple.edu/~hbling/data/TColor-128/TColor-128.html}}, OTB2015\footnote{\label{fn:1}The sequences together with the ground-truth and matlab code is available at: \url{http://cvlab.hanyang.ac.kr/tracker_benchmark/datasets.html}} and VOT2015\footnote{\url{http://www.votchallenge.net/vot2015/dataset.html}} and their related evaluation protocols \cite{Authors31,Authors32,Authors53} to evaluate
the proposed ScaleSSKCF algorithms. First, we introduce the experimental setup. Next, we evaluate two variants of our proposed method, i.e., OWSC (our algorithm without structural constraint) and OWTC (our algorithm without temporal consistent), to analyze the effect of structural constrain term and temporal constrain term in our proposed method. Finally, our proposed algorithm is compared with some the most related state-of-the-art methods.
\subsection{Experimental Setup}
\begin{figure}
  \centering
  \includegraphics[width=8cm,height=3cm]{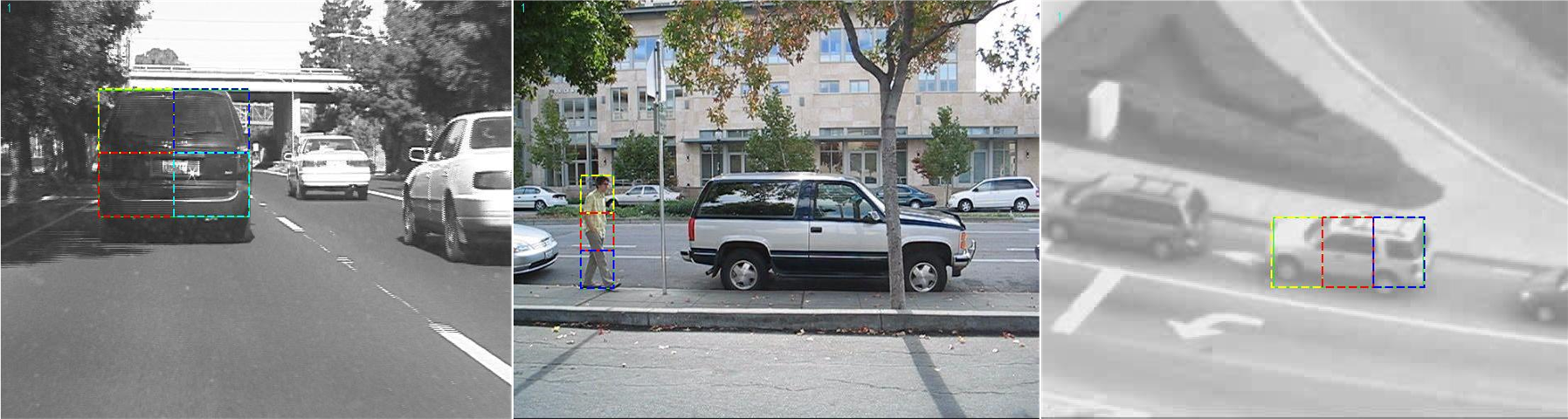}\\
  \caption{Visualization of the target's partition based on the target's aspect ratio}\label{fig1}
\end{figure}

Our proposed approach is implemented in native MATLAB 2014a on a 3.6GHZ Intel i7 Core4 PC with 4G RAM. The average running speed is around 40 frames per second. The optimization takes 5 iterations in the first frame and 2 or 3 iterations for each online update. In our method, the feature extraction takes up 48\% of the total consuming time. But the optimization is only 3\%.\\
\textbf{Parameters:} Our tracker involves a few model parameters, \emph{i.e.}, trade-off parameter $C$, scale parameter $\eta$ and shape parameter $\lambda$ of confidence maps, the weight parameter $\delta$ of the Laplacian regulation term, the controlling factor $\beta$ of the temporal constrain term, and lower and upper thresholds $(\theta_{l},\theta_{u})$ in (\ref{eq7}). In addition, other parameters include the smooth factor $\kappa$ in (\ref{eq20}) and hyperparameter in (\ref{eq40}). For online tracking, the model is updated by linear interpolation with the adaption rate $\varrho$ in (\ref{eq44}). In our experiments, the detailed parameters setting is shown in Table \ref{Tb:1}, where padding means
the magnification of the image region samples relative to the target bounding box.
\begin{table*}
\caption{Parameters setting of our proposed method (ScaleSSKCF).}
\begin{center}
\small
\setlength{\tabcolsep}{0.5mm}{
\begin{tabular}{ccccccccccccccc}
  \toprule
parameters & padding &$\eta$ & $(\theta_{l},\theta_{u})$ & C &$\lambda$&$\delta$&$\beta$&$\kappa$&$\epsilon$&$\varepsilon$&$\varrho$&bins of HOG& cell size&orientations\\
\hline
  Value &1.8&$0.1*\sqrt{MN}$&$(0.4,0.9)$&$10^{4}$&2 &0.05 &5 &3 &5.5 &0.2 &0.015&31 & $4\times 4$&9 \\
  \bottomrule
\end{tabular}}
\end{center}
\label{Tb:1}
\end{table*}
The number of local parts $L$ is adaptively determined by the aspect ratio of object $\frac{O_{N}}{O_{M}}$, where $O_{N}$ and $O_{M}$ separately denote the width and height of object. If $0.6 <\frac{O_{N}}{O_{M}}< 1.6$, the target is divided into $2\times 2$ local parts, \emph{i.e.}, $L=4$; if $\frac{O_{N}}{O_{M}}\leq 0.6$, the target is partitioned into $3\times 1$ local parts, \emph{i.e.}, $L=3$; if $\frac{O_{N}}{O_{M}}\geq 1.6$, we sample $1\times 3$ local parts on the target. The detailed partitioning method is shown in Fig.\ref{fig1}. Note that, any other part sampling methods can also be adopted.
\\
\textbf{Datasets and Evaluation Metrics}: To assess the performance of the proposed tracker, extensive experiments are carried on several public benchmark datasets such as TempleColor128 \cite{Authors53}, OTB-2015 \cite{Authors52} and VOT2015 \cite{Authors32}. In the TempleColor128 and OTB-2015 datasets, we adopt two metrics used in \cite{Authors53} including distance precision (DP) and overlap precision (OP). The DP is the relative number of frames in the sequence where the center location error is smaller than a certain threshold. As in \cite{Authors31}, the DP values at a threshold of 20 pixels are reported in our work. The OP is defined as the percentage of frames where the bounding box overlap surpasses a certain threshold. We report the results at a threshold of 0.5, which correspond to the PASCAL evaluation criterion \cite{Authors61}. Except for the DP and OP metrics, the precision and success plots \cite{Authors31} have also been adopted to measure the overall tracking performance. For the precision and success plots, we respectively use the DP value of each tracker and the area under curve (AUC) score of each success plot to rank the tracking algorithms.
In VOT2015 sequences, we utilize evaluation criterion proposed in \cite{Authors32}.
\subsection{Key Component Validation}
Here, on the TempleColor128 dataset \cite{Authors53}, we discuss the impact of structural constraint term and temporal consistent term in our algorithm. Based on the algorithm analysis in Section \ref{algsection}, the performance of our algorithm should decrease to some extent without structural constraint term and temporal consistent term, which is shown in Table \ref{Tb:2}. The OWSC and OWTC respectively denote the absence of structural constraint term and temporal consistent term in our model. Overall, the performance of the proposed algorithm is best among these three methods (\emph{e.g.} OWSC, OWTC and ScaleSSKCF (ours)). Seen from the comparison, the performance of OWSC is worst, which means that the structural constraint term of our tracking model plays the most important role in the performance of our algorithm.
\begin{table}[!h]
\caption{Comparing the results of OWSC, OWTC and ScaleSSKCF based on mean distance precision (DP) and mean overlap precision (OP). The entries in \textcolor[rgb]{1.00,0.00,0.00}{red} denote the best results.}
\begin{center}
\begin{tabular}{cccc}
  \toprule
  Metrics & OWSC & OWTC & ScaleSSKCF (Ours) \\
  \hline
  mean OP (\%) & 45.3 &47.1 &\textcolor[rgb]{1.00,0.00,0.00}{\textbf{47.7}} \\
  mean DP (\%)  & 60.9  &62.9 &\textcolor[rgb]{1.00,0.00,0.00}{\textbf{64.1}} \\
  \bottomrule
\end{tabular}
\end{center}
\label{Tb:2}
\end{table}

\subsection{Evaluation on OTB2015 dataset}

Here, we provide a comparison of our method with 7 state-of-the-art and the most related methods from the literature: SRDCF \cite{Authors26}, RPT \cite{Authors46}, SKSCF \cite{Authors45}, samf \cite{Authors24}, Staple \cite{Authors35}, lct2 \cite{Authors33} and DPCF \cite{Authors51} on the OTB2015 dataset. But a few most related methods (\emph{e.g.}, SCF \cite{Authors48} and RPAC \cite{Authors47}) are not included in our comparative experiments because their source codes are not open to the public and they didn't do the corresponding experiments on this dataset in their paper.
\subsubsection{State-of-the-art Comparison}
The quantitative comparison among these selected methods is reported in Table \ref{Tb:OTB2015-OPE}, using mean overlap precision (OP) and mean distance precision (DP) over all 100 video sequences of OTB2015. Seen from the Table \ref{Tb:OTB2015-OPE}, our method achieves the best result by 72\% on the mean OP metric. However, the SRDCF obtains the best result on the mean DP.  The main reason is that the SRDCF introduces the spatial regulation term to deal with the boundary effect caused by the FFT, which makes it learn a more discriminative model. The performance of our method is almost the same as its but our speed is about 20 times faster (For a more fair comparison of speed, please refer to the results in VOT2015). Although the performance of DPCF is slightly superior to our method on the mean DP, the score of its OP is obviously lower than the one of our method, that's because the DPCF limits the scale changing range of the target between 0.75 and 1.25 and can't estimate it accurately when the target occurs the large-scale change.
\begin{table*}
\caption{Comparison with state-of-the-art trackers on the 100 sequences of OTB2015. The top two results are highlighted by bold and different colors: \textcolor[rgb]{1.00,0.00,0.00}{red} and \textcolor[rgb]{0.00,0.07,1.00}{blue} color.}
\begin{center}
\setlength{\tabcolsep}{0.5mm}{
\begin{tabular}{ccccccccc}
  \toprule
  Metrics &RPT& SKSCF & DPCF &SRDCF &samf &lct2 & Staple&ScaleSSKCF (Ours) \\
  \hline
  mean OP (\%) &63.2 &67.0&68.9&\textcolor[rgb]{0.00,0.07,1.00}{\textbf{71.6}}&67.9&63.3 &70.5&\textcolor[rgb]{1.00,0.00,0.00}{\textbf{72.0}} \\
  mean DP (\%) &76.0 &78.1&77.8&\textcolor[rgb]{1.00,0.00,0.00}{\textbf{78.8}}&77.0 &77.0&\textcolor[rgb]{0.00,0.00,1.00}{\textbf{78.5}}&77.6 \\
  mean FPS (s)&1.8&\textcolor[rgb]{0.00,0.07,1.00}{\textbf{23}}&20&2 &9 &6 &12&\textcolor[rgb]{1.00,0.00,0.00}{\textbf{41}}\\
  \bottomrule
\end{tabular}}
\end{center}
\label{Tb:OTB2015-OPE}
\end{table*}
\begin{figure}[t]
   \centering
   \subfloat[]{\label{Fig:DP-OPE}
   \includegraphics[width=0.5\linewidth]{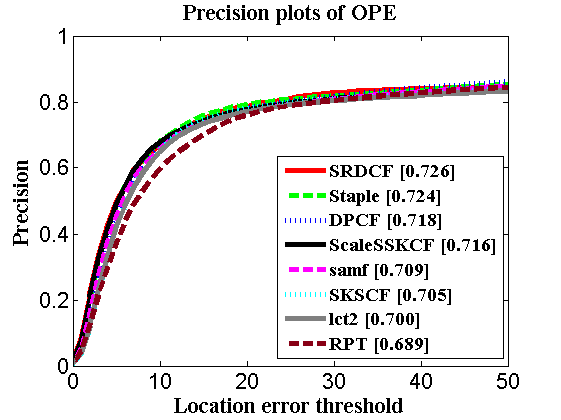}}
   \subfloat[]{\label{Fig:OP-OPE}
   \includegraphics[width=0.5\linewidth]{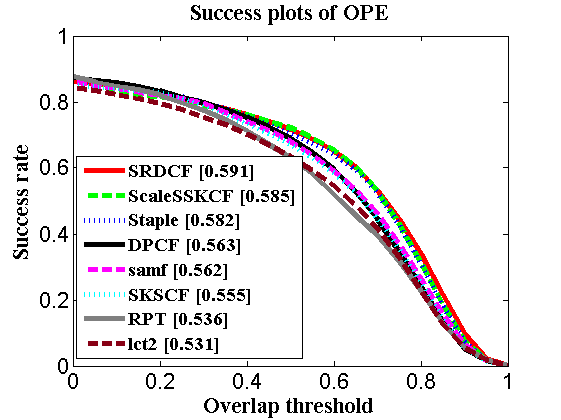}}\\

   \caption{Precision and success plots over all 100 sequences in OTB2015. The area under the curve (AUC) scores of each tracker are reported in the legends.}
\label{fig:OTB2015-OPE}
\end{figure}

Fig. \ref{fig:OTB2015-OPE} gives precision and success plots over all 100 sequences in OTB2015. The success plot shows the ratios of successful frames at the thresholds varied from 0 to 1. While the precision plot describes the ratios of frames in which the center location error (CLE) is smaller than a arbitrary threshold ranging from 0 to 50 pixels. The trackers of each sub-figure in Fig. \ref{fig:OTB2015-OPE} are ranked by their area under the curve (AUC) scores, displayed in the legend. In the success plots of OPE, our method shows comparable results as the SRDCF and significantly outperforms other several correlation filter trackers. For the precision plots, our method is slightly inferior to the DPCF by 0.2\%, that may be because the DPCF combines the tracking results of global correlation filter model, and our method also inferior to the Staple by 0.8\%, the main reason of which is that the color histogram model of Staple is more robust to the deformation of target.
\subsubsection{Attribute Based Comparison}
\begin{figure*}[!h]
\includegraphics[width=13.5cm,height=11.0cm]{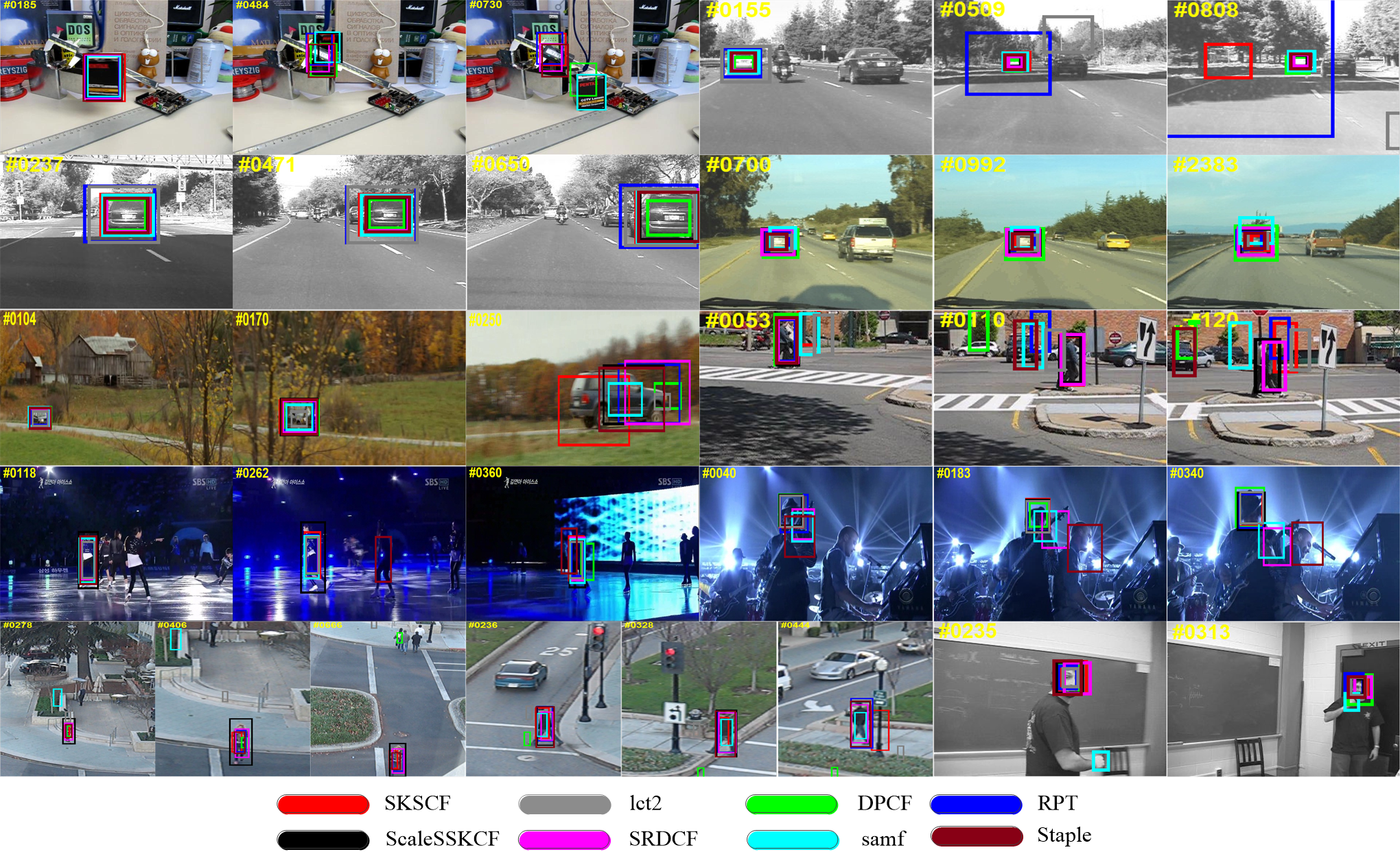}
   \caption{Qualitative comparison of our approach with 7 state-of the-art trackers (denoted in different colors) on the several typically challenging sequences (from left to right and top to down are Box, Car1, Car4, Car24, CarScale, Couple, Skating1, Shaking, Human5, Human6 and Freeman1 respectively).}
\label{fig6}
\end{figure*}

The sequences in OTB2015 are annotated with 11 different attributes to describe the different challenges in the tracking problem, including illumination variation (IV), scale variation (SV), occlusion (OCC), deformation (DEF), motion blur (MB), fast motion (FM), in plane rotation (IPR), out-of-plane rotation (OPR), out-of-view (OV), background clutters (BC), and low resolution (LR). These attributes are useful for analyzing the performance of trackers in different aspects. The Tables \ref{Tb:OTB2015-OPE-OP} and \ref{Tb:OTB2015-OPE-DP} respectively shows the performance of ours and 7 state-of-the-art methods in terms of AUC (success metrics) and DP (precision metrics) with respect to each attribute . In Table \ref{Tb:OTB2015-OPE-OP}, our method has gained 7 the best and 2 the second best out of 11 subcategories for AUC score. In case of deformation, compared with other methods, our method achieves the second best results (The DP score on the center location error is 73.4\% and the AUC score on the overlap rate is 54.6\%), which is inferior to the ones of Staple because the color histogram model used in the Staple is more robust to the deformation of target. As are shown in Tables \ref{Tb:OTB2015-OPE-OP} and \ref{Tb:OTB2015-OPE-DP}, for the sequences involving the fast motion, the performance of our method and other part-based trackers become bad because the searching area of part-based tracking method shrinks, leading to drift problem. However, the SRDCF can obtain the best results because it can learn a strong model that adapts the fast motion of target on the larger samples. For scale variation, our method achieves the better results than other methods except the SRDCF. Note that the DPCF adopts the scale estimation technique similar to ours but its performance is significantly inferior to ours (\emph{e.g.,}, our AUC score on the overlap rate exceeds it by 5\%). That is because the DPCF limits the scale changes in a small range (from 0.75 to 1.25), which let it not adapt to the large scale change of the target. For the occlusion factor, our tracker obtains the best AUC score of 58.5\% on the overlap rate. The main reason is that our method eliminates the effect of the occlusion when updating the discriminative model. For the low resolution sequences, our method obtains the best results which may be attribute to the temporal consistent term in our model.
\begin{table*}
\caption{Success metrics (\%) of the trackers for 11 attributes. The top two results are highlighted by  \textcolor[rgb]{1.00,0.00,0.00}{red} and \textcolor[rgb]{0.00,0.07,1.00}{blue}.}
\begin{center}
\setlength{\tabcolsep}{0.8mm}{
\begin{tabular}{cccccccccccc}
  \toprule
  Attributes &FM& BC & MB &DEF&IV&IPR & LR&OCC&OPR&OV&SV \\
  \hline
  RPT &52.0 &58.1&51.3&50.2 &53.9&52.5&36.2&48.2&50.9&42.2&48.1 \\
  SKSCF &52.6 &58.0&52.0&51.0&55.5 &\textcolor[rgb]{0.00,0.07,1.00}{54.4}&35.1&52.1&52.8&40.1&48.8 \\
  DPCF&50.3&\textcolor[rgb]{0.00,0.07,1.00}{58.3}&53.2&52.9&57.5 &52.4 &40.0&54.4&\textcolor[rgb]{0.00,0.07,1.00}{55.2}&43.5&50.5\\
  SRDCF&\textcolor[rgb]{1.00,0.00,0.00}{58.4} &56.6 &\textcolor[rgb]{1.00,0.00,0.00}{58.3} &53.1 &58.7&52.3&\textcolor[rgb]{0.00,0.07,1.00}{49.1}&\textcolor[rgb]{0.00,0.07,1.00}{56.2}&53.8&43.9&\textcolor[rgb]{1.00,0.00,0.00}{55.6}\\
  samf&53.0 &55.3 &53.3 &50.9 &54.8&53.1&42.8&55.3 &54.2&\textcolor[rgb]{0.00,0.07,1.00}{48.9}&50.7\\
  lct2&50.5 &54.1 &51.2 &49.5 &51.9&53.0&29.9&48.9&51.2&42.9&43.3 \\
  Staple&\textcolor[rgb]{0.00,0.07,1.00}{53.9}&57.3&\textcolor[rgb]{0.00,0.07,1.00}{53.9}&\textcolor[rgb]{1.00,0.00,0.00}{56.3}&\textcolor[rgb]{0.00,0.00,1.00}{59.1}&54.3&39.9&55.4&54.1&46.5&52.7\\
  ScaleSSKCF(our)&52.8 &\textcolor[rgb]{1.00,0.00,0.00}{63.3} &52.0 &\textcolor[rgb]{0.00,0.00,1.00}{54.6} &\textcolor[rgb]{1.00,0.00,0.00}{60.6}&\textcolor[rgb]{1.00,0.00,0.00}{55.0}&\textcolor[rgb]{1.00,0.00,0.00}{50.7}&\textcolor[rgb]{1.00,0.00,0.00}{58.5}&\textcolor[rgb]{1.00,0.00,0.00}{57.0}&\textcolor[rgb]{1.00,0.00,0.00}{51.9}&\textcolor[rgb]{0.00,0.07,1.00}{55.5}\\
  \bottomrule
\end{tabular}}
\end{center}
\label{Tb:OTB2015-OPE-OP}
\end{table*}
\begin{table*}
\caption{Precision metrics (\%) of the trackers for 11 attributes. The top two results are highlighted by  \textcolor[rgb]{1.00,0.00,0.00}{red} and \textcolor[rgb]{0.00,0.07,1.00}{blue}.}
\begin{center}
\setlength{\tabcolsep}{0.8mm}{
\begin{tabular}{cccccccccccc}
  \toprule
  Attributes &FM& BC & MB &DEF&IV&IPR & LR&OCC&OPR&OV&SV \\
  \hline
  RPT &68.5 &\textcolor[rgb]{0.00,0.07,1.00}{81.4}&68.1&72.3 &\textcolor[rgb]{1.00,0.00,0.00}{80.1}&74.4&59.5&68.0&72.7&54.5&71.2 \\
  SKSCF &\textcolor[rgb]{0.00,0.00,1.00}{70.7} &\textcolor[rgb]{1.00,0.00,0.00}{82.1}&67.9&69.7&76.8 &\textcolor[rgb]{0.00,0.07,1.00}{77.7}&63.0&72.3&76.0&55.0&72.3 \\
  DPCF&67.6&79.8&68.7&\textcolor[rgb]{0.00,0.00,1.00}{73.4}&\textcolor[rgb]{0.00,0.07,1.00}{79.2} &74.4 &66.7&73.2&\textcolor[rgb]{1.00,0.07,0.00}{76.4}&54.5&71.7\\
  SRDCF&\textcolor[rgb]{1.00,0.00,0.00}{74.3} &74.5 &\textcolor[rgb]{1.00,0.00,0.00}{73.5} &72.8 &76.1&72.1&65.9&74.2&74.4&57.3&\textcolor[rgb]{1.00,0.00,0.00}{75.4}\\
  samf&69.9 &74.3 &67.6 &69.0 &74.2&74.1&\textcolor[rgb]{0.00,0.07,1.00}{68.4}&\textcolor[rgb]{1.00,0.00,0.00}{75.3} &75.8&\textcolor[rgb]{1.00,0.07,0.00}{66.3}&73.5\\
  lct2&68.0 &75.9 &66.8 &70.6 &74.6&\textcolor[rgb]{1.00,0.00,0.00}{78.2}&53.7&69.7&75.9&58.2&69.1 \\
  Staple&70.9&77.4&\textcolor[rgb]{0.00,0.07,1.00}{69.8}&\textcolor[rgb]{1.00,0.00,0.00}{76.8}&78.2&76.8&61.0&74.3&74.9&65.8&73.8\\
  ScaleSSKCF(our)&68.5 &\textcolor[rgb]{1.00,0.00,0.00}{82.1} &68.2 &\textcolor[rgb]{0.00,0.00,1.00}{73.4} &77.3&75.2&\textcolor[rgb]{1.00,0.00,0.00}{70.7}&\textcolor[rgb]{0.00,0.07,1.00}{74.9}&\textcolor[rgb]{0.00,0.00,1.00}{76.3}&\textcolor[rgb]{0.00,0.00,1.00}{66.0}&\textcolor[rgb]{0.00,0.07,1.00}{74.5}\\
  \bottomrule
\end{tabular}}
\end{center}
\label{Tb:OTB2015-OPE-DP}
\end{table*}
Fig. \ref{fig6} shows a qualitative comparison of our approach with 7 existing methods on 11 challenging example videos. Both the SRDCF and ScaleSSKCF perform well in the presence of heavy occlusion (\emph{e.g.}, Human6), which can be attributed to the fact that SRDCF learns a discriminative model on larger image region and our method removes the effects of heavy occlusion when updating our tracking model. The lct2 can effectively re-detect target in the case of tracking failure, e.g., the sequence with the heavy occlusion (Shaking), but it can not perform well in scale variation and illumination changes (\emph{e.g.}, Car1, Car4 and Car24). When the tracked target of the sequence is occluded by similar color barrier (\emph{e.g.}, Box), the Staple performs very bad because the color histogram model used by it does not distinguish them. Compared with these method, our method can estimate the object size more accurately when occurring the large scale variation (\emph{e.g.}, Car1, Car4, Car24, CarScale and Human5). For the sequences (e.g., Skating1) including the deformation, our approach significantly outperforms other several methods because it adopts some tricks (\emph{e.g.}, structural constraint term and temporal consistency term ) to make our discriminative model more robust for target deformation.
\subsubsection{Robustness Evaluation to Initialization}
\begin{figure}[t]
   \centering
   \subfloat[]{\label{Fig:DP-OPE}
   \includegraphics[width=0.5\linewidth]{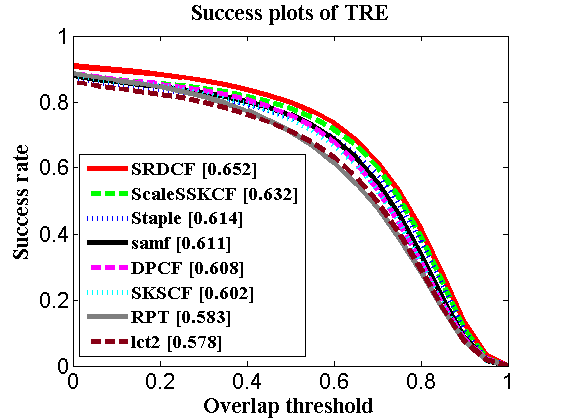}}
   \subfloat[]{\label{Fig:OP-OPE}
   \includegraphics[width=0.5\linewidth]{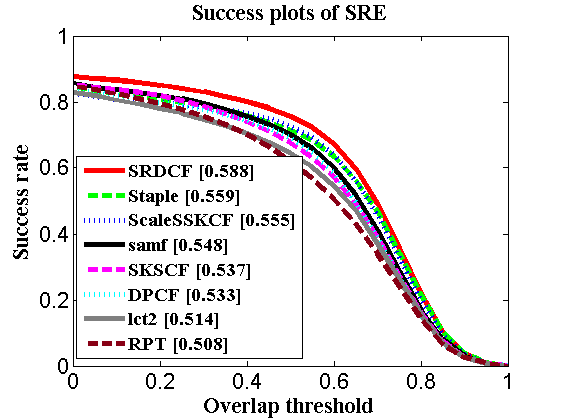}}\\

   \caption{An evaluation of the temporal and spatial robustness to initializations on the OTB2015 dataset. The area under the curve (AUC) scores of each tracker are reported in the legends.}
\label{fig:OTB2015-TSRE}
\end{figure}

We adopt two robustness metrics: spatial robustness (SRE) and temporal robustness (TRE) provided by \cite{Authors31} to evaluate the robustness to initializations. The SRE criteria initializes the tracker with perturbed boxes, which the TRE criteria starts the tracker at the frame corresponding to each segmentation point (each sequence is divided into the 20 segmentation points). The Fig. \ref{fig:OTB2015-TSRE} shows the TRE and SRE success plots of ours method compared with other related trackers. In the success plots of TRE, the performance of our method is second only to that of the SRDCF but is significantly superior to the rest of trackers, especially DPCF and RPT. For the SRE criteria, our method also is slightly inferior to the Staple by only 0.4\%. This evaluation demonstrates our method is relatively robust to different spatial and temporal initializations.
\subsection{Evaluation on TempleColor 128 dataset}
\begin{figure}
  \centering
  \includegraphics[width=14cm,height=6cm]{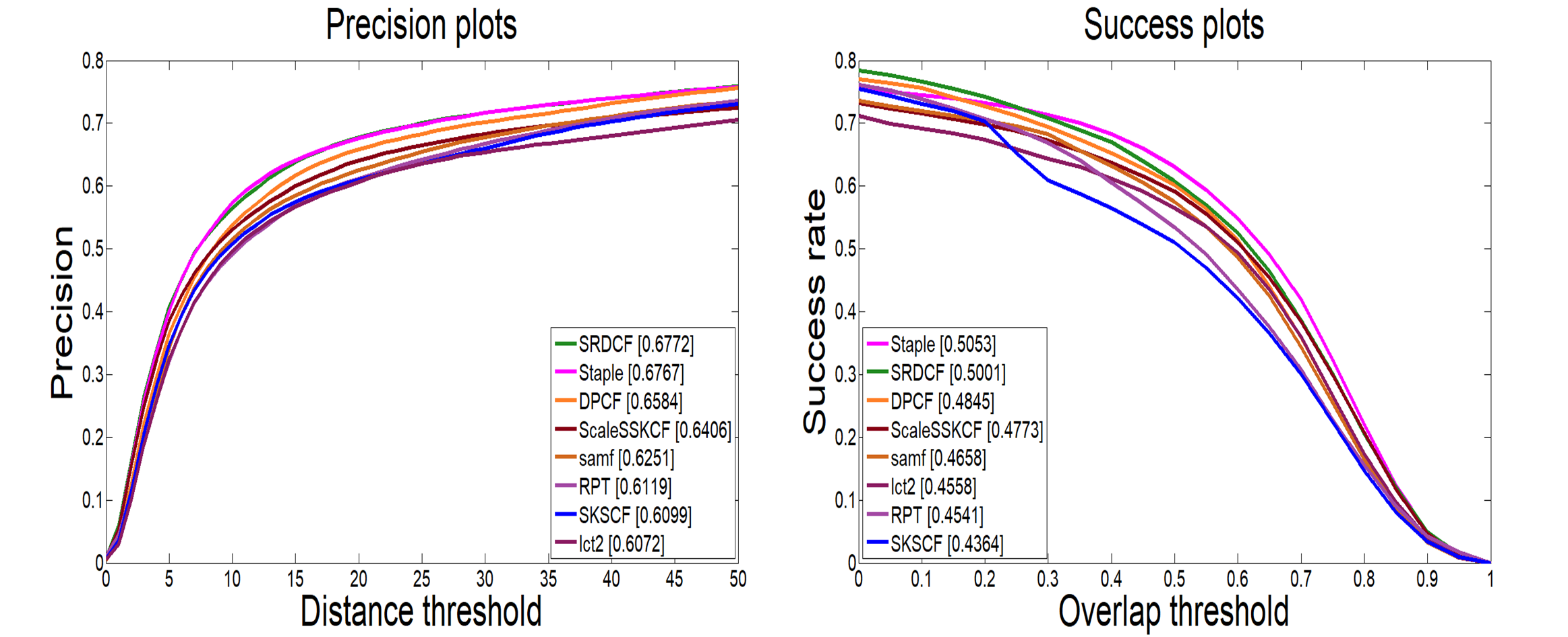}\\
  \caption{Precision and success plots over all 128 sequences in TempleColor128 dataset. For the success plots, the area under the curve (AUC) scores of each tracker are reported in the legends. And the precision obtained at threshold 20 is shown in the legends of the precision plots.}\label{Tcolor128}
\end{figure}

Here, we evaluate our method on the TempleColor128 dataset. The Fig.\ref{Tcolor128} shows a comparison with 7 state-of-the-art and the most related methods from the literature: Staple \cite{Authors35}, SRDCF \cite{Authors26}, DPCF \cite{Authors51}, lct2 \cite{Authors33}, SKSCF \cite{Authors45}, RPT \cite{Authors46} and samf \cite{Authors24}. The performance of our method is only ranked the fourth in these methods. The main reason is that TempleColor128 dataset contains about a half sequences with the fast motion \cite{Authors53} and our method is not suitable for dealing with these sequences because the valid searching region becomes smaller when the target is divided into the patches. For the SRDCF, it can learn the filter from the larger searching region because of the spatial regularization term, which makes it against the fast motion. However, its speed is only about 2 frame per second and it can not meet the realtime applications. The DPCF integrates the results of local and global correlation filter by the minimum spanning tree model. Although its performance is slightly superior to our method, the complex model brings down its speed.
\subsection{Evaluation on VOT2015}
Finally, we compare our method with other 9 related trackers (CCOT \cite{danelljan2016beyond}, deepSRDCF, Staple \cite{Authors35}, SRDCF \cite{Authors26}, DPT\cite{Authors56}, samf \cite{Authors24}, DPCF \cite{Authors51}, SKSCF \cite{Authors45} and lct2 \cite{Authors33}) on VOT2015 consisting of 60 challenging videos.
Here, we evaluate the performance of the trackers by three metrics (accuracy (overlap with ground truth), robustness (failure rate) and excepted average overlap (EAOP)) provided in \cite{Authors32}. In VOT2015, a tracker is restarted in the case of a failure. In more detail, we refer the readers to \cite{Authors32}. The Table \ref{Tb:VOT2015} gives their comparison results on VOT2015 according to three metrics mentioned above. Among the compared methods, our method is only ranked the fourth. Note that the results of CCOT and deepSRDCF directly come from the VOT2016 competition. The CCOT and deepSRDCF both use the more discriminative deep convolution features. According to the conclusions in \cite{wang2015understanding}, the better features can dramatically improve the tracking performance than the tracker its. Thus, it is not fair to directly compare our method and them.
\begin{table*}
\caption{Comparison results on the VOT2015 dataset. The top two results are highlighted by  \textcolor[rgb]{1.00,0.00,0.00}{red} and \textcolor[rgb]{0.00,0.07,1.00}{blue}.}
\begin{center}
\small
\setlength{\tabcolsep}{0.5mm}{
\begin{tabular}{ccccccccccc}
\toprule
  &CCOT&deepSRDCF&Staple&SRDCF&DPT&samf&DPCF&SKSCF&lct2&ScaleSSKCF(ours)\\
\hline
Accuracy&0.52&\textcolor[rgb]{1.00,0.00,0.00}{0.56}&0.53&0.53 &0.48 &0.51&0.51&0.50&0.52&\textcolor[rgb]{0.00,0.07,1.00}{0.55}\\
Robustness&\textcolor[rgb]{1.00,0.00,0.00}{0.85}&\textcolor[rgb]{0.00,0.07,1.00}{1.00}&1.35&1.53&1.75&2.08&2.15&2.40&2.52&1.75\\
EAOP&\textcolor[rgb]{1.00,0.00,0.00}{0.325}&\textcolor[rgb]{0.00,0.07,1.00}{0.318}&0.291&0.245&0.234&0.202&0.191&0.185&0.175&0.252 \\
\bottomrule
\end{tabular}}
\end{center}
\label{Tb:VOT2015}
\end{table*}

\begin{figure}
  \centering
  \includegraphics[width=12cm,height=7cm]{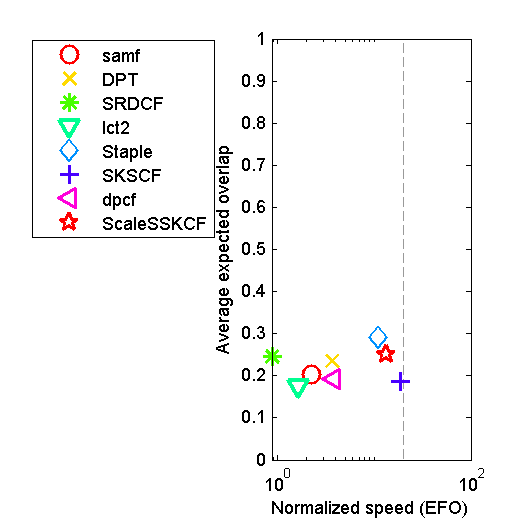}\\
  \caption{Expected average overlap scores with respect to the tracking speed in
EFO units. The dashed vertical line denotes the estimated real-time performance
threshold of 20 EFO units.}
\label{VOT2015-speed}
\end{figure}

As is known, except the accuracy and robustness, the tracking speed is also very crucial in many real tracking application. Therefore, we visualize the expected overlap score with respect to the tracking speed measured in EFO units in Fig. \ref{VOT2015-speed}, where we exclude the CCOT and deepSRDCF for fairness. Seen from the Fig. \ref{VOT2015-speed}, our method achieves the better balance between the performance and speed.
\section{Conclusions}
In this paper, we proposed a scale-adaptive structural support kernel correlation filter tracking model, which is called ScaleSSKCF. Our method combines part-based tracking strategy into support correlation filter tracker by the structural constraint term of the proposed model, which remains the strong discriminability of the support correlation filter (SCF) and also preserves the spatial structure of the target. To reduce the issues of drifting away from the object, we consider the temporal consistency of each part in our model. In addition, we also introduce the occlusion detection and scale estimation into the proposed tracking method, which makes our tracker less sensitive to some complex factors (\emph{e.g.}, partial occlusion and scale variation ). Results on three benchmark datasets show that our tracker performs favorably against several state-of-the art tracking methods in terms of accuracy, robustness and speed.
\section{Acknowledgment}
This work is supported by the National Natural Science Foundation of China
Under Grant No. 61602288, 61703252 and Shanxi Provincial Natural Science Foundation of China Under Grant No. 201701D221102. The authors also would like to thank the anonymous reviewers for their valuable suggestions.




\bibliographystyle{model3-num-names}
\bibliography{egbib}







\end{document}